\pdfoutput=1
\documentclass[conference]{IEEEtran}

\usepackage{indentfirst}
\usepackage{preprint}
\PassOptionsToPackage{hyphens}{url}\usepackage{hyperref}
\usepackage{graphicx}
\usepackage{adjustbox}
\usepackage[bottom]{footmisc}
\usepackage[hyphens]{url}
\usepackage{hyperref}
\usepackage[hyphenbreaks]{breakurl}

\usepackage{amsmath, amsthm, amssymb, amsfonts}

\usepackage[numbers]{natbib}
\usepackage[utf8]{inputenc}	% allow utf-8 input
\usepackage[T1]{fontenc}	% use 8-bit T1 fonts
\usepackage{xcolor}		% colors for hyperlinks
\usepackage{booktabs} 		% professional-quality tables
\usepackage{nicefrac}		% compact symbols for 1/2, etc.
\usepackage{microtype}		% microtypography
\usepackage{lineno}		% Line numbers
\usepackage{float}			% Allows for figures within multicol
\usepackage{newfloat}
\DeclareFloatingEnvironment[name={Supplementary Figure}]{suppfigure}
\usepackage{sidecap}
\sidecaptionvpos{figure}{c}

\usepackage{titlesec}

\usepackage{authblk}

\begin{document}

\title{A SMART RECYCLING BIN USING WASTE IMAGE CLASSIFICATION AT THE EDGE}

% Option 2 for author list
\author{\begin{tabular}[t]{c@{\extracolsep{15em}}c} 
Xueying Li and Ryan Grammenos \\
candy.li.19@ucl.ac.uk   ,  r.grammenos@ucl.ac.uk \\
Electrical and Electronic Engineering   \\ 
University College London  \\
\end{tabular}
}

\twocolumn[ % Method A for two-column formatting
 % Method A for two-column formatting
  
\maketitle
\begin{abstract}
\textbf{Rapid economic growth gives rise to the urgent demand for a more efficient waste recycling system. This work thereby developed an innovative recycling bin that automatically separates urban waste to increase the recycling rate. We collected 1800 recycling waste images and combined them with an existing public dataset to train classification models for two embedded systems, Jetson Nano and K210, targeting different markets. The model reached an accuracy of 95.98\% on Jetson Nano and 96.64\% on K210. A bin program was designed to collect feedback from users. On Jetson Nano, the overall power consumption of the application was reduced by 30\% from the previous work to 4.7 W, while the second system, K210, only needed 0.89 W of power to operate. In summary, our work demonstrated a fully functional prototype of an energy-saving, high-accuracy smart recycling bin, which can be commercialized in the future to improve urban waste recycling.} 

\end{abstract}
%\keywords{First keyword \and Second keyword \and More} % (optional)
\vspace{0.35cm}

 % Method A for two-column formatting
] % Method A for two-column formatting

%%%%%%%%%%%%%%%  Main text   %%%%%%%%%%%%%%%
% \linenumbers
\section{Introduction}
Waste generation has increased dramatically in the 21st century due to the growth in the global population. According to the World Bank Group's report \cite{WorldBankGroup2018What2.0}, 2.01 billion tonnes of municipal solid waste are generated worldwide every year, with only about 19\% adequately recycled  \cite{WorldBankGroup2018What2.0}. Recycling not only helps to preserve raw material but, more importantly, reduces the landfills required, which is an undesirable way of waste disposal due to its high demand for space and the danger of introducing contaminants into the ground or even groundwater system. A major step of recycling is to separate the waste into specific categories according to its material. Failure to do so will significantly harm the effectiveness of recycling. Traditionally, workers at the recycling company sort the waste into corresponding categories by hand, which is an inefficient method and requires unnecessary labor. Therefore, people are now in dire need of a more advanced and automated waste separation system.\par
Image classification machine learning (ML) algorithms have been used to build automatic waste classification systems and assist waste management. This research aimed to improve the existing classification system and pursue the commercialization of an artificial intelligence (AI) street bin. Specifically, we focused on reducing the power consumption of the controller board to extend AI bin’s battery life, increasing its waste classification accuracy and reducing its price. \par
The only intelligent recycling bin on the market, Bin-e  \cite{Bin-e}, uses image classification algorithms to separate the trash into four categories: plastic, paper, metal and glass. It achieves a waste segmentation accuracy of 92\% \cite{Bin-e}. However, the price USD 5200 \cite{Bin-e} is expensive for a rubbish bin. Also, Bin-e’s operation requires a 230V power supply, so it cannot replace the traditional trash bins at the locations without plug sockets. Therefore, it can hardly increase the household and street recycling rate. As a result, this paper proposes low-cost battery-powered real-time recycling waste segmentation bin systems to fill this gap.\par
In this research, we trained a light MobileNet image classification model for Jetson Nano with TrashNet  \cite{GaryThungGarythung/trashnet:Classification}, an open-source waste dataset, and 1800 new training samples collected by us. The model demonstrated great performance with a high test accuracy of 95.98\% and a small parameter size of 3.0 M. Jetson Nano only consumed 4.7 W when it ran the model. We also found a cheaper and less energy-hungry device, K210 \cite{CanaanKendryteK210} to further reduce power consumption. The K210 board only consumed 0.89 W at inference time and the model on it achieved a high test accuracy of 96.64\%, which made it a more practical solution for trash bin applications. The code used in this study is available on Github \cite{XueyingLiMolvcanAN-SMART-RECYCLING-BIN-USING-WASTE-IMAGE-CLASSIFICATION-AT-THE-EDGE}.\par
This paper is organized as follows. In Section \ref{sec:2}, the related works performed in waste classification and AI trash bins are introduced. Section \ref{sec:3} introduces the system design and explains the background theories that support this study. The methodology of data collection and model training is introduced in section \ref{sec:4}. Section \ref{sec:5} analyses the test results and evaluates the system performance on Jetson Nano and K210. Finally, Section \ref{sec:6} concludes the achievements and provides suggestions for future research.

\section{Related works}\label{sec:2}

The United Kingdom (UK) government has planned to increase the household recycling rate in England to 50\% by 2020, but only 44\% of municipal waste was reused and recycled in 2020 \cite{2022Progress2020}. The inconvenience and lack of knowledge are two crucial factors that prevent people from recycling \cite{Knickmeyer2020SocialAreas}. Automatic waste segmentation bins were designed to overcome these problems by helping people classify and send the waste into the corresponding containers, making waste disposal more convenient. Table \ref{tab:1} outlines the approaches and the primary hardware components involved in related studies to construct the waste segmentation systems. 

\begin{table*}[!ht]
 \caption{Related works in AI bins}
  \centering
  \begin{adjustbox}{width=1\textwidth}
      \small
\begin{tabular}{p{0.7cm}p{2cm}p{4cm}p{4cm}p{4cm}}
\toprule
\cmidrule(r){1-5}
Year &	Author &	Waste classification method &	Classification category &	The device \\
\midrule
2014    & Rajkamal et   al. \cite{Rajkamal2014AGREENBIN}  &Inductive   metal sensor,\newline capacitive based moisture sensor, methane sensor, odor sensor & metal/glass, food,   bio,\newline paper/plastic, inert                         & PIC18F4580   \newline controller             \\
2017                                               & kumar et al. \cite{B.R.S.Kumar2017Eco-FriendlyManagement}                       & Inductive   and capacitive metal sensor, gas sensor, bacteria sensor, moisture sensor   & Biodegradable,   plastic, metal   & STM32   controller  \\
2019                                               & Pereira et al. \cite{Pereira2019SmartOptimisation}                          & Infrared radiation   sensor, capacitive sensor                                          & Dry, wet, plastic                                                      & Atmega 328P \newline microcontroller         \\
2019                                               & Ziouzios et al. \cite{Ziouzios2019AClassification}                          & CNN image   classfication                                                               & Glass, paper,   metal, plastic, cardboard, trash                       & Xilinx Pynq-Zl FPGA                 \\
2020                                               & White et al. \cite{White2020WasteNet:Bins}                             & CNN image   classification                                                              & Paper, plastic,   metal, glass,\newline cardboard, other                       & Jetson Nano                         \\
2021                                               & Jimeno et al. \cite{Jimeno2021DevelopmentProcessing}                            & CNN   image classification                                                              & Aluminum   cans, plastic bottles, plastic cups, paper, spoons/forks    & Computer   with \newline Nvidia RTX 2060 GPU \\
2021\newline                                               & Sallang et al. \cite{Sallang2021AEnvironment}                           & CNN   image classification \newline                                                             & Glass, paper,   metal, plastic, cardboard                              & Raspberry   Pi 4 \newline         \\
\bottomrule
\end{tabular}
  \end{adjustbox}
  \label{tab:1}
\end{table*}

In early research, a microcontroller is connected to a variety of sensors to determine the composition of the waste. For example, inductive and capacitive sensors can detect the metal element \cite{Rajkamal2014AGREENBIN}\cite{B.R.S.Kumar2017Eco-FriendlyManagement}\cite{Pereira2019SmartOptimisation} and moisture sensors separate wet waste from dry waste  \cite{B.R.S.Kumar2017Eco-FriendlyManagement}\cite{Pereira2019SmartOptimisation}. The microcontroller makes decisions based on its readings \cite{Rajkamal2014AGREENBIN}\cite{B.R.S.Kumar2017Eco-FriendlyManagement}. However, although the sensor-based classification method can detect the composition precisely, it lacks the ability to classify waste into more specific groups. The sensors cannot discriminate between plastic, paper, glass, and other unrecyclable dry waste, which are important categories in recycling. \par

The development of machine learning and image classification enables the bin to sort the waste based on visual input like a human. The convolutional neural network (CNN) is a branch of image classification algorithms that performs mainly convolution operations on the pixels \cite{Yamashita2018ConvolutionalRadiology}. It is the most popular choice due to its high accuracy and power efficiency compared to other methods and is used in all the four papers \cite{Ziouzios2019AClassification}\cite{White2020WasteNet:Bins}\cite{Jimeno2021DevelopmentProcessing}\cite{Sallang2021AEnvironment}. It gives the bin ability to differentiate between spoons and cups, which is tremendous progress compared to the sensor-based approaches. Traditionally, the CNN models run on a cloud raising the data transmission latency and user privacy security problems. To solve these problems, recent research moved the computation to an edge embedded system. However, edge computing has the drawback of limiting computation resources, so the model size is important in selecting the CNN model structure.\par

To develop, evaluate and select the proper CNN structures, most research in this field used the TrashNet dataset developed by Mindy Yang and Gary Thung \cite{GaryThungGarythung/trashnet:Classification} in 2017, the first open-access recycling waste dataset. This high-quality dataset contains 2527 waste photos from 6 groups: metal, glass, plastic, paper, cardboard, and trash. It provides a foundation for later research and our study also used this dataset. \par

Efforts have been made to increase the segmentation accuracy of CNN models based on TrashNet. The state-of-art accuracy of 97.86\% on TrashNet was reached in \cite{OzkayaFine-TuningRecyclability}. However, the CNN architecture, GoogLeNet, used in this research will cause out-of-memory (OOM) on Jetson Nano \cite{Abdulmahmood2021ImprovingAnalysis}. It demonstrated the potential of CNN classification, but the model size and computation cost must be cut down to implement the machine learning algorithm on embedded systems. A lighter CNN model, WasteNet, was constructed by White et al. and achieved an accuracy of 97\% on the TrashNet dataset \cite{White2020WasteNet:Bins}. The paper claimed that the model could be loaded to Jetson Nano, but it did not provide details regarding the edge implementation, such as the classification speed. Nevertheless, from the model structure, we could estimate that the classification speed for one image would be too slow for real-time classification. It can only be used in a bin application that classifies the waste objects based on one photo. \par

The classification speed is quantified by inference time, which refers to the time taken for one classification to be completed. Our application looks for a smaller model that can interact with the user in real-time, with an inference time that must be smaller than 0.1 s, which is the human visual reaction time. The CNN model, EfficientNet B0, used in \cite{Abdulmahmood2021ImprovingAnalysis} is the starting point of this research and it achieved an accuracy of 95.38\% and an inference time of 0.07 s on Jetson Nano. While the model is fast enough to be used in real-time applications, the 96\% high memory usage needs optimization for bin applications. \par

Power consumption is another factor that needs to be considered in the final product. Unfortunately, rare research has paid attention to it. For instance, none of the trash bin cases listed in Table \ref{tab:1} has measured the power consumption of the proposed system. Still, it is evident that the CNN-based approaches have higher power consumption than the sensor-based one and the Pynq-Zl and Raspberry Pi will typically have lower power consumption than Jetson Nano used in \cite{White2020WasteNet:Bins} and \cite{Abdulmahmood2021ImprovingAnalysis}. The Raspberry Pi 4 has a typical power consumption of between 2.7 W and 6.4 W. It reduces the power consumption but results in undesirable low performance \cite{Sallang2021AEnvironment}. As a result, the CNN architecture, MobileNet V2, has a low average per-class precision of 91.76\% and a long inference time of 0.358 s on it. \par

The limitation of the current research is that the energy-saving systems will have unacceptable low performance in the classification task for commercialization. As a result, this study designed two high-accuracy waste classification systems with lower power consumption than previous studies. The first system reduced the power consumption of the application on Jetson Nano by using a lighter model, MobileNet, to reduce power consumption while maintaining the accuracy. \par

The second system is developed based on K210, a less expensive and more energy-saving embedded device.  K210 is an unpopular choice and is used in fewer than 100 research. Most research implemented YOLO object detection models on it \cite{Henderi2020AnArea}\cite{Li2021ADRC-basedFeedback} and demonstrated its outstanding power efficiency. K210 has not been used for recycling waste classification when this paper was written, so this paper proposed and evaluated an innovative approach.

\section{System design}\label{sec:3}
\subsection{The AI recycling bin design }

The AI bin consists of five recycling waste containers and a detection box. The waste will be sorted after it is placed in the detection box. The whole system can be controlled by a Jetson Nano or a K210 board. \par

\begin{figure}
        \includegraphics[width=0.5\textwidth]{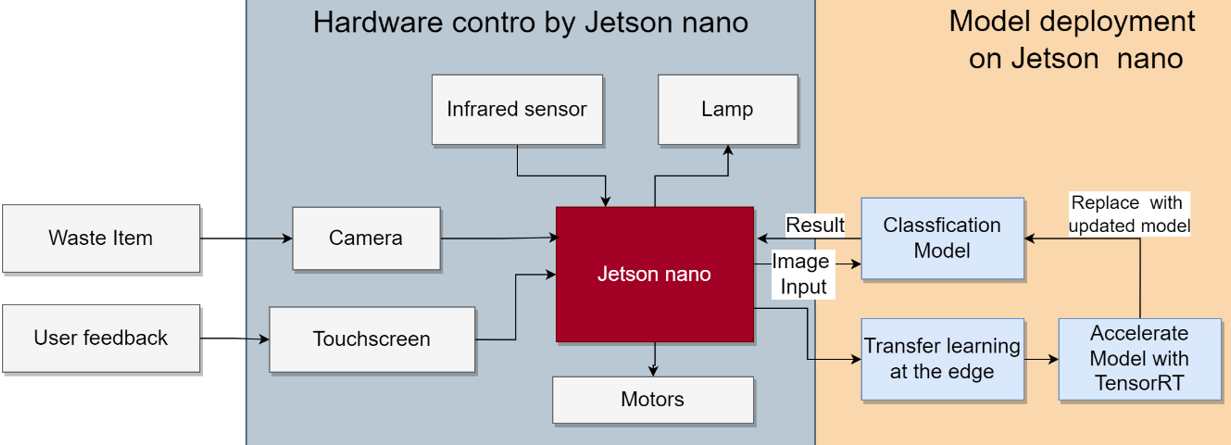}
        \caption{System block design of WasteNet.}
        \label{fig:1}
\end{figure}

The bin design using Jetson Nano is summarized in Figure \ref{fig:1}. Jetson Nano will interact with the users and collects feedback through the touch screen in front of it, displaying the instructions for using the bin and the camera inputs. The Raspberry Pi camera at the top takes photos of the waste in the detection box. The images will be fed to the classification algorithm in Jetson Nano, which classifies the photos into seven groups. 
The waste classification models in the previous study \cite{Abdulmahmood2021ImprovingAnalysis} have five output classes: "paper", "metal", "plastic", "cardboard" and "glass". They will be reproduced in the next section as the benchmark models of our study. Our model will be trained with two new classes, "empty" and "hand". "Empty" means there is nothing in the photo, while the "hand" group detects a human's hand to avoid trapping the user's hand by the door. When these two groups are detected, the bin waits and continues detecting. 
\subsection{The Benchmark models }\label{bench}
The waste classification models in \cite{Abdulmahmood2021ImprovingAnalysis} consist of an input layer, a pre-processing augmentation layer, an EfficientNet B0 base model layer, a global average pooling 2D layer and a dense layer. The first model has an input layer size of 512x384 pixels, while the second model input has 384x288 pixels. Table \ref{tab:2} shows the configuration of the augmentation layer and the model training hyperparameters for reproduction. \par
\begin{table}
  \centering
    \small
    \caption{Training parameters used in \cite{Abdulmahmood2021ImprovingAnalysis} }
  \begin{tabular}{ll}
    \toprule
    \cmidrule(r){1-2}
    Pre-Processing Method     & Value \\
    \midrule
    Random flip & Horizontal and vertical \\
    Random rotation & Up to 180\(^{\circ}\) \\
    Random translation & Up to 10\% \\
    Random zoom & Up to 75\% \\
    \bottomrule
    \cmidrule(r){1-2}
    Training Parameter     & Value \\
    \midrule
    Learning rate scheduler & Constant learning rate scheduler \\
    Train/Validation/Test split ratio & 72/18/10\\
    Optimizer & Adam optimiser \\
    Training epochs & 50\\
    Learning rate&	4.3e-05\\
    Fine-tuning training epochs	&8\\
Fine-tuning learning rate&	4e-06\\
Loss functions&	Sparse categorical cross entropy\\
Classifier activation function	&Softmax\\
Include top layers in base model&	False\\
    Batch size & 16 \\
    \bottomrule
  \end{tabular}
  \label{tab:2}
\end{table}
The models were trained with the TrashNet dataset, which consists of six categories of images: glass, paper, cardboard, plastic, metal, and trash. Only the first five groups of data were used to build the model that classified recycling into five categories. In total, 2152 images were used for training and validation, and 238 images were used for testing. The train/validation/test split is 72/18/10.\par

\begin{figure}[H]
 \centering
        \includegraphics[width=0.4\textwidth]{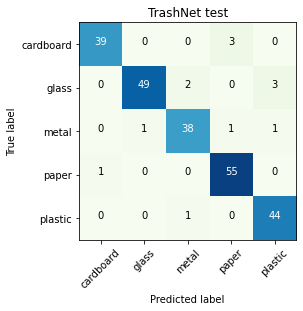}
        \caption{Confusion matrix of the benchmark models}
        \label{fig:2}
\end{figure}

To begin with, the base model is initialized with pre-trained weights on ImageNet. The two models are trained separately by setting the corresponding input sizes. The learning rate was set to 4.3e-05 at the first 50 epochs and was reduced to 4e-06 at the eight epochs afterwards. Eventually, both models achieved the same test accuracy of 95.38\%. Figure \ref{fig:2} shows the confusion matrix of the test results and indicates that most mistakes are made in classifying between paper and cardboard.\par
After the model was reproduced, it was saved into TensorFlow SavedModel format and transferred to Jetson Nano. The 384x288 model was optimized successfully through TensorRT on Jetson Nano, but the 512x384 model caused an OOM error. The throughput of the accelerated model was 19 inferences per second (IPS) at the maximum power mode.\par
Nevertheless, the 384x288 model occupied 94.9\% of memory. The system would freeze up if we ran other applications simultaneously. To implement the classification model in an AI bin application, it is essential to replace the original base model with the lighter model, such as MobileNet V3, which has lower memory usage and power consumption. 
\subsection{MobileNet V3 model}
The EfficientNet architecture used in the benchmark model has demonstrated outstanding performance on the ImageNet dataset. In particular, EfficientNet B7 has the same top-1 accuracy, 84.3\%, as the GPipe while being 8.4x \cite{Howard2017MobileNets:Applications}. However, as EfficientNet B0 is the smallest model in the family, with a parameter size of 4.0 M without top fully connected layers, so MobileNet V3 Large, which only has 3.0M parameters and top-1 accuracy of 1.1\% less than EfficientNet on ImageNet, is chosen as the substitute. 
MobileNet is a series of CNN architectures developed by Google. It contains three architectures: MobileNet V1 \cite{Howard2017MobileNets:Applications}, V2 \cite{Sandler2018MobileNetV2:Bottlenecks} and V3 \cite{Howard2019SearchingMobileNetV3}. MobileNet V2 achieved higher accuracy on ImageNet with a smaller parameter size than MobileNet V1. The MobileNet V3 optimized the latency of MobileNet V2. \par

Other CNN models also achieved similar accuracies as benchmark models, such as ShuffleNet V2, ResNet50 and Dense-169 as listed in Table \ref{tab:3}, but their parameter size is too large and results in high latency. The parameter size of the MobileNet V3 Large is smaller than EfficientNet B0 when the fully connected layers are removed in both models.

\begin{table}[H]
  \centering
    \small
    \caption{Model classification accuracy on ImageNet  }
  \begin{tabular}{p{3cm}p{2cm}p{2.5cm}}
    \toprule
    \cmidrule(r){1-3}
    CNN   architecture          & Number of   \newline parameters (M) & Top-1   accuracy on ImageNet (\%) \\
    \midrule
EfficientNet   B0 \cite{Tan2019EfficientNet:Networks}  & 5.3                        & 76.3                              \\
MobileNet V3 Large\newline  \cite{Howard2019SearchingMobileNetV3}& 5.4                        & 75.2                              \\
MobileNet V1 \cite{Howard2017MobileNets:Applications}       & 4.2                        & 70.6                              \\
ShuffleNet V2 \cite{MaShuffleNetDesign}     & 7.4                        & 74.9                              \\
Dense-169 \cite{Huang2016DenselyNetworks}         & 14.3                       & 76.2                             \\
ResNet50 \cite{He2015DeepRecognition}          & 25. 6                        & 74.9                       \\
    \bottomrule
    
  \end{tabular}
  \label{tab:3}
\end{table}

\subsection{Jetson Nano and Kendryte K210 chip}
The model is deployed to Jetson Nano in the previous study \cite{Abdulmahmood2021ImprovingAnalysis}. Jetson Nano \cite{NvidiaJetsonKit} is a powerful single-board computer designed by Nvidia for AI applications. It runs the Linux embedded system, which can execute multiple applications and graphics processing unit (GPU) accelerated neural networks in parallel. Its developer kit has a high-definition multimedia interface (HDMI) port, two camera serial interface (CSI) camera slots and four universal serial bus (UCB) ports that can connect to accessories. The 40-pin header allows it to drive motors and receive signals from sensors. It has a 5 W mode and a maximum power mode with typical power consumption between 5 W and 10W. \par
In this study, we also explored the use of a K210 developer kit to run the waste classification model. K210, in contrast, only runs one model stored in its flash memory during operation without an operating system. Canaan developed it to perform tasks specialized in image and audio processing. It has a neural network processor called knowledge processor unit (KPU) used to accelerate CNN computations, such as convolution, batch normalization, activation, and pooling operations. Images can be read from a digital video port (DVP) camera and displayed on a liquid-crystal display (LCD) screen connected to board. The Sipeed M1w dock K210 developer kit has a maximum power consumption of 3 W. \par
Table \ref{tab:4} shows the specifications of the two embedded devices. Jetson Nano has greater flexibility and computation power for applications, while K210 has lower price and power consumption. K210 only supports 1x1 and 3x3 kernels in CNN, so it cannot implement most CNN structures such as MobileNet V3. The low random-access memory (RAM) limits the model size to 6 MB and the input resolution to 320x240 pixels. In contrast, the 4GB RAM allows Jetson Nano to run more complex models such as Efficient Net B0 with an input of 512x384 pixels. Therefore, K210 is a cheaper substitute for Jetson Nano which runs models lighter than MobileNet V1.\par 

\begin{table}[H]
  \centering
    \small
    \caption{Comparison of Jetson Nano and K210 }
  \begin{tabular}{p{3cm}p{2cm}p{2.5cm}}
    \toprule
    \cmidrule(r){1-3}
                       & Nvidia Jetson   Nano 4GB\cite{NvidiaJetsonKit} & Sipeed M1w   dock K210 kit\cite{2019SipeedFeatures} \\
    \midrule

Clock                  & 1.43GHz                  & 0.4GHz                     \\
RAM                    & 4GB                      & 8 MB                       \\
Power consumption      & 5-10W                    & 0.3W-3W   (typically 1W)   \\
Programming   language & Python                   & Micropython                \\
Cost                   & GBP 133                  & GBP 23 \\                 
    \bottomrule
    
  \end{tabular}
  \label{tab:4}
\end{table}

After the models are trained in TensorFlow, they are optimized for on-device machine learning before deployment to reduce the size and latency. TensorFlow Lite library \cite{TensorFlowTensorflow.org/lite/performance/model_optimization} can compress the TensorFlow model through quantization, which reduces the precision of the weights from 32-bit floats to 16-bit floats or 8-bit integers. K210 uses nncase \cite{KendryteKendryteNncase}, an open-source library developed by Kendryte, to accelerate the TensorFlow Lite model into its unique kmodel format that optimizes KPU usage through constant-folding, operator replacement and operator fusion. Firstly, constant-folding computes the expressions that only contain known constants and substitutes the answers to save computation cost at runtime. Then, operator replacement changes some operators in the TensorFlow Lite model, which KPU cannot accelerate, with several operators that perform a similar task. Finally, operator fusion combines several operators into a single KPU accelerated operator. These operations optimizes the latency of model running on K210.\par
For Jetson Nano, Nvidia's TensorRT \cite{NVDIANVIDIATensorRT}  library is used to optimize inference on its GPU and provides post-training quantization for TensorFlow model.

\section{Methodology}\label{sec:4}

This section introduces the development of the classification model on Jetson Nano and K210. We collected 1872 new recycling waste data to train new models. Experiments were carried out to evaluate the models' accuracy and power consumption.\par
\subsection{Collect more training data}
Firstly, more recycling waste images were collected to increase the training dataset. As the TrashNet dataset only contains waste images with white background, making the classification model vulnerable to variation in lighting conditions and dirt or contamination in the background. Waste photos with a black background were taken to tell the model the background color is irrelevant in classification and increase the robustness of real-world cases. Additionally, we want to add two classes, "empty" and "hand" to the model, but we do not have images belonging to these categories to train the model. Therefore, we built a detection model with plywood material and collected a domain-specific waste dataset.\par
\begin{figure}[H]
 \centering
        \includegraphics[width=0.3\textwidth]{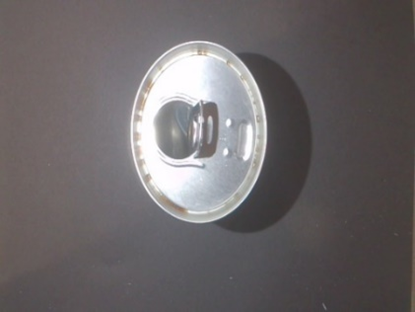}
        \caption{A sample waste image taken by the cameras}
        \label{fig:3}
\end{figure}

The waste samples were collected from a nearby recycling bin and from people directly. Figure \ref{fig:3} shows a waste photo taken by the camera at the top.\par
\begin{figure}[H]
 \centering
        \includegraphics[width=0.3\textwidth]{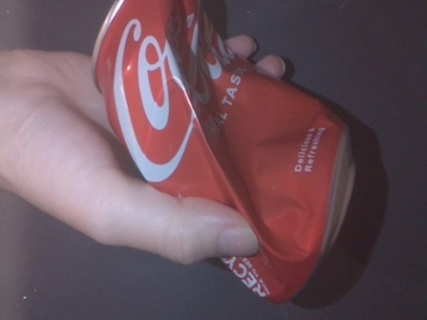}
        \caption{Sample "hand" images taken by the top camera}
        \label{fig:4}
\end{figure}

The "hand" group contains photos of the waste held by hand. Figure \ref{fig:4} shows an example of the "hand" class. If there is a hand in the images, it belongs to the "hand" group, regardless of the type of waste present.\par
\begin{figure}[H]
 \centering
        \includegraphics[width=0.3\textwidth]{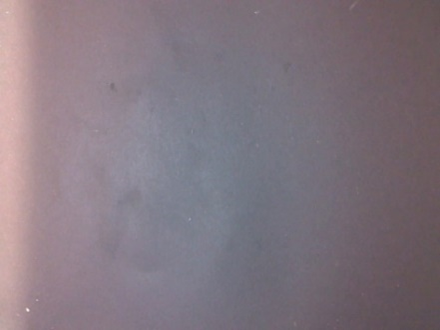}
        \caption{"Empty" images taken by the top camera}
        \label{fig:5}
\end{figure}

The "empty" class only contains a black background. This class tells the bin application that there is nothing in the detection box. An example is shown in Figure \ref{fig:5}.
We collected 1871 images and combined them with the TrashNet dataset. Table \ref{tab:5} lists the number of images in each class and each dataset.

\begin{table}[H]
  \centering
    \small
    \caption{Number of waste items in each category}
  \begin{tabular}{p{1.5cm}p{3cm}p{3cm}}
    \toprule
    \cmidrule(r){1-3}
 Category  & Number of images in TrashNet & Number of images in our dataset \\
    \midrule

Cardboard & 403                          & 268                             \\
Glass     & 501                          & 130                             \\
Paper     & 594                          & 89                              \\
Plastic   & 482                          & 298                             \\
Metal     & 410                          & 140                             \\
Hand      & 0                            & 920                             \\
Empty     & 0                            & 26               \\                 
    \bottomrule
    
  \end{tabular}
  \label{tab:5}
\end{table}

Although recycling waste was divided into five categories, some of the waste was a mixed waste that contained more than one group of materials. For example, a coca-cola tin had a plastic label wrapped around the metal body and a wine glass bottle had a metal cap. In this case, we labelled the items according to the largest component of them. Additionally, different countries and recycling companies have different recycling rules. In the UK, the paperboard materials, such as cartons, are recycled with paper, while in the United States of America (USA), they belong to cardboard. In this paper, we follow the recycling guide of a waste management company in the USA \cite{WMIntellectualPropertyHoldings2020Recycling-Myths}. When it comes to commercialization in the future, it is crucial to consider the recycling rules of the local community.\par

\subsection{Train the models with transfer learning}
The two datasets were used to train new classification models with TensorFlow version 2.9. Firstly, we replaced the EfficientNet B0 model layer in the benchmark model with MobileNet V3 architecture initialized with the pre-trained weights on ImageNet. The MobileNet model has a smaller parameter size and memory usage of the model. Then, the output size of the dense layer was changed to seven to add two new classes to the model. Finally, a resizing layer is inserted between the data augmentation layer and the base model layer to resize the input from 512x384 pixels to 224x224 pixels. By compressing the input after data augmentation, we obtained a clearer image for base model input than compressing the data before the data augmentation layer. \par
\begin{figure}[H]
 \centering
        \includegraphics[width=0.5\textwidth]{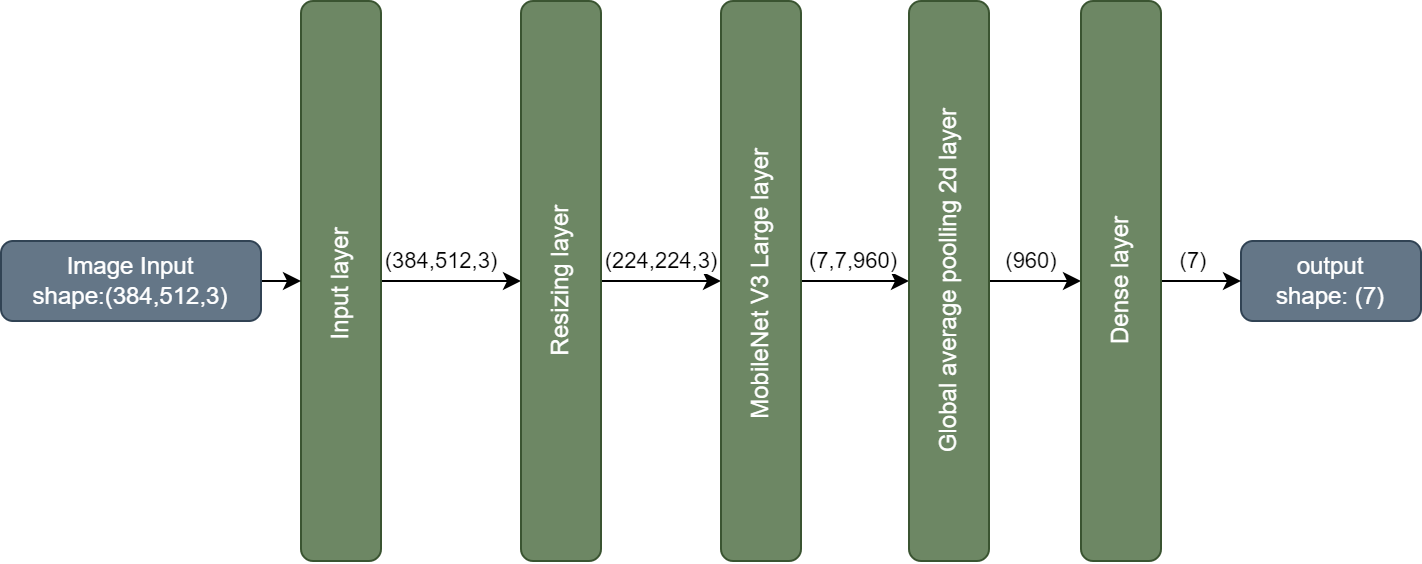}
        \caption{Waste classification model structure}
        \label{fig:6}
\end{figure}

Figure \ref{fig:6} illustrates the structure of our model. The model was trained with the hyperparameters shown in Table \ref{tab:6}. The final model was converted to the TensorRT model with TensorFlow version 2.5, and its performance was measured on Jetson Nano. 

\begin{table}[H]
  \centering
    \small
    \caption{Training parameters used to build the classification model on Jetson Nano}
  \begin{tabular}{ll}
    \toprule
    \cmidrule(r){1-2}
    Training Parameter     & Value \\
    \midrule
    Learning rate scheduler & Constant learning rate scheduler \\
    Train/Validation/Test split ratio & 72/18/10\\
    Optimizer & Adam optimiser \\
    Training epochs & 50\\
    Learning rate&	4.3e-05\\
    Fine-tuning training epochs	&10\\
Fine-tuning learning rate&	4e-06\\
Loss functions&	Sparse categorical cross entropy\\
Classifier activation function	&Softmax\\
Include top layers in base model&	False\\
    Batch size & 16 \\
    Base model dropout rate& 0.2 \\
    \bottomrule
  \end{tabular}
  \label{tab:6}
\end{table}
\subsection{Build and implement the K210 model}
\begin{figure}
 \centering
        \includegraphics[width=0.5\textwidth]{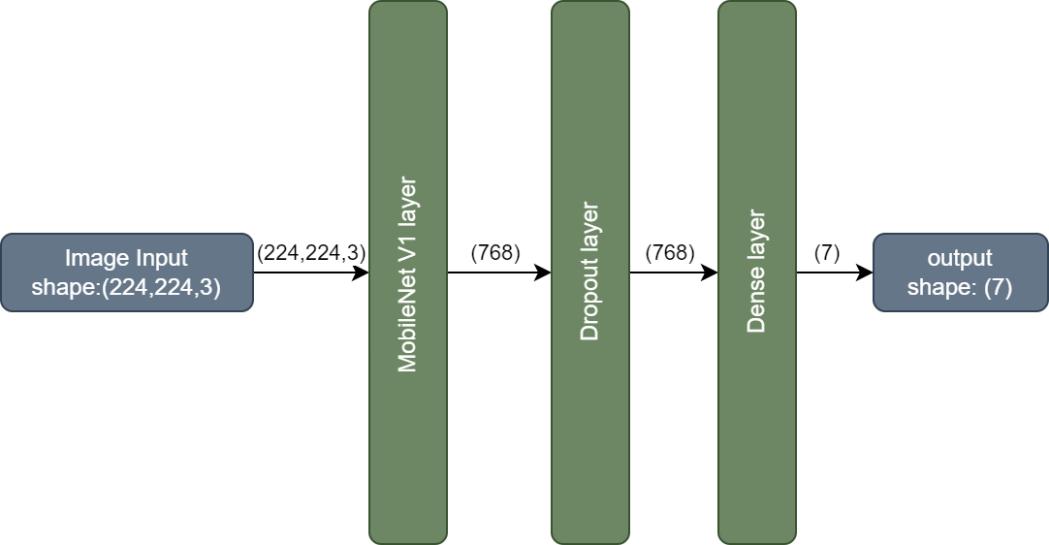}
        \caption{MobileNet V1 structure used by K210}
        \label{fig:7}
\end{figure}

After building the AI application on Jetson Nano, we used a cheaper and less power-hungry processor, K210, to run the waste classification model. K210 used its own model format called kmodel. Sipeed provides sample code to train classification model based on MobileNet V1 and convert it to kmodel model format with nncase \cite{SipeedSipeedMaix_train}. They applied data augmentation to the image data, as shown in Table \ref{tab:7}, and built a model with structure with TensorFlow, shown in Figure \ref{fig:7}. 
\begin{table}[H]
  \centering
    \small
    \caption{Data augmentation used in \cite{SipeedSipeedMaix_train}}
  \begin{tabular}{ll}
    \toprule
    \cmidrule(r){1-2}
    Pre-Processing Method     & Value \\
    \midrule
    Random width and height shift& Up to 20\% \\
    Random rotation & Up to 180\(^{\circ}\) \\
    Random translation & Up to 10\% \\
    Random shear & Up to 50\% \\
    \bottomrule
  \end{tabular}
  \label{tab:7}
\end{table}

We kept the same data augmentation setup and used the same training hyperparameters applied in the previous training, as shown in Table \ref{tab:8}. All training data is resized into 224x224 pixels before training. The model used the same test/validation split 70/15 as our previous models on Jetson Nano. We first trained 110 epochs to find the stopping epochs and then trained the model with the stopping epochs. 
\begin{table}[H]
  \centering
    \small
    \caption{Training parameters used to build K210 model }
  \begin{tabular}{p{4cm}p{4cm}}
    \toprule
    \cmidrule(r){1-2}
    Training Parameter     & Value \\
    \midrule
    Learning rate scheduler & Constant learning rate scheduler \\
    Optimizer & Adam optimiser \\
    Learning rate&	4.3e-05\\
Loss functions&	Sparse categorical cross entropy\\
Classifier activation function	&Softmax\\
Include top layers in base model&	False\\
    Batch size & 16 \\
Dropout rate in base model&	0.001\\
Dropout rate at output&	0.001\\
    \bottomrule
  \end{tabular}
  \label{tab:8}
\end{table}
We tested both the implementation of MobileNet V1 and V2 on K210. K210 only supported MobileNet V1 with alpha equal to 0.75 and MobileNet V2 with alpha equal to 0.5. OOM error occurs when a larger model is deployed. The ImageNet  accuracies of the V1 (alpha = 0.75) and V2 (alpha = 0 .5) were 68.4\% \cite{Tensorflow/models/research/slim/nets/mobilenet_v1.md} and 65.4\% \cite{Tensorflow/models/research/slim/nets/mobilenet/} correspondingly, so only MobileNet V1 architecture was trained.\par
The trained model is accelerated into kmodel format and loaded onto K210 together with the firmware using Kflash provided by Kendryte \cite{KendryteKendryteKflash.py}. We used the firmware that only supports basic Application Programming Interface (API) and integrated development environment (IDE) on K210 to reduce memory usage.

\subsection{Measure the performance of the model on the devices}
   \begin{figure}
 \centering
        \includegraphics[width=0.4\textwidth]{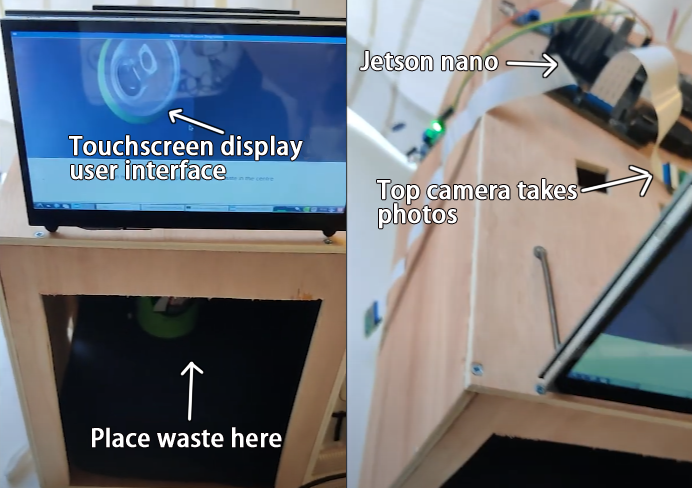}
        \caption{Photo of the final waste classification device}
        \label{fig:75}
\end{figure}
After the classification model was trained, the throughput of each model on Jetson Nano was measured by recording the latency. Figure \ref{fig:75} demonstrated the final waste detection device. The memory usage and power consumption of the device were measured when only the model was running and when the AI application was running at the same time. The memory usage was recorded from the system tool of Jetson Nano. We used Keweisi voltage current meter to measure the power. It was connected between the power supply and the device to test the power consumption of Jetson Nano and K210.

\section{Results and discussion}\label{sec:5}
The accuracies of models obtained from transfer learning will be analyzed in this section. Our best model, the 224x224 pixels MobileNet V3 Large model, achieved an accuracy of 95.98\% in classifying the images into seven groups. The model had 40 IPS on Jetson Nano at 5 W power mode and consumed 30\% less power than the EfficientNet B0 model. The K210 model achieved a further 96.64\% on waste classification while the board's power only drew 0.89 W. 
\subsection{Accuracies of models trained with TrashNet dataset on Jetson Nano }

As mentioned in section \ref{bench}, The EfficientNet B0 model has achieved high top-1 accuracy on original test data. However, we could not use this compute-intensive model directly in the application. So, we aimed to construct a less resource-intensive model while maintaining higher accuracy than the benchmark model.\par

Firstly, we applied a 70/15/15 training/validation/test split to the TrashNet dataset by randomly separating 15\% of TrashNet data to form the TrashNet test set. The rest of the TrashNet dataset became a new TrashNet training set. This ratio was used because the author in \cite{YuTowardsClassification} demonstrated that both 15\% and 10\% could produce the same level of classification accuracy on the TrashNet dataset, which is higher than 20\% and 25\% with statistical analysis. Other research \cite{Aral2019ClassificationModels}\cite{Bircanoglu2018RecycleNet:Networks} also utilized 70\% of the TrashNet data as training data. So, we consider that 70/15/15 would also be suitable for model training.\par

We trained and tested EfficientNetB0, MobileNet V3 Large and MobileNet V3 Small with new TrashNet training set using the hyperparameters in Table \ref{tab:6}. All the models converged quickly and reached 90\% of the highest validation accuracy after 8 epochs. Table \ref{tab:9} compares the results we obtained.

\begin{table*}[!ht]
 \caption{Accuracy of three models on 70/15/15 split}
  \centering
  \begin{adjustbox}{width=1\textwidth}
      \small
\begin{tabular}{p{3cm}p{2cm}p{2.5cm}p{2.5cm}p{2.5cm}p{2.5cm}}
\toprule
\cmidrule(r){1-6}
Model \newline Architecture & Total \newline parameters (M) & Total keras(.h5)\newline model size (MB) & Training \newline accuracy (\%) & Top-1 Validation \newline Accuracy (\%) & Top-1 Accuracy \newline on test set (\%) \\
\midrule
EfficientNetB0     & 4.2                  & 15.8                               & 99.44                  & 96.32                          & 94.40                           \\
MobileNet V3 Large & 3.0                  & 11.7                               & 99.50                  & 94.94                          & 95.52                           \\
MobileNet V3 Small & 0.94                 & 3.87                               & 98.94                  & 93.33                          & 93.56                          \\
\bottomrule
\end{tabular}
  \end{adjustbox}
  \label{tab:9}
\end{table*}
As it can be observed, the MobileNet V3 Small model had lower training, validation, and test accuracy than the other two models. In contrast, MobileNet V3 large reached similar training accuracy as EfficientNet B0. The test accuracy was 1.12\% higher, while the validation accuracy was 1.38\% lower. The two models had similar performance on this task while MobileNet V3 Large was lighter and faster, so MobileNet V3 Large was used to build our new model. \par

As the original TrashNet test set only has five classes, 15\% of the "hand" and "empty" group data were separated from our new dataset and combined with the TrashNet test set to form the seven-class test set, which evaluated the model's performance on the two new class. The rest of our dataset was added to TrashNet training data and formed the final training data.\par

We trained EfficientNetB0 and MobileNet V3 Large with the new training data. The same hyperparameters in Table \ref{tab:6} were used to train both models. 
The result is shown in Table \ref{tab:10}. The accuracy on 5-class test set of the EfficientNet B0 model increased from 94.40\% in Table \ref{tab:9} to 95.23\%. This indicates that the new dataset had a positive effect on the models and helped them to learn more general features that assist classification tasks. Our photos have different waste samples, lighting and background, so the accuracy on TrashNet will not increase significantly. \par

\begin{table*}[!ht]
 \caption{Accuracies of models trained by the final training data}
  \centering
  \begin{adjustbox}{width=1\textwidth}
      \small
\begin{tabular}{p{3cm}p{2cm}p{1.2cm}p{2cm}p{2.4cm}p{2cm}p{2.4cm}}
\toprule
\cmidrule(r){1-7}
Model \newline Architecture & Base model \newline input resolution (pixels) & Training accuracy (\%) & Top-1 Accuracy on 5-class test set (\%) & Per-class precision of 5-class test set (\%) & Top-1 Accuracy \newline on new 7-class test set (\%) & Per-class precision of 7-class test set (\%) \\
\midrule
EfficientNet B0    & 512x384                                                                           & 99.59                  & 95.23                                   & 94.79                                        & 95.98                                       & 95.97                                        \\
MobileNet V3 Large & 512x384                                                                           & 99.46                  & 94.68                                   & 94.53                                        & 95.38                                       & 95.68                                        \\
MobileNet V3 Large & 224x224                                                                           & 99.46                  & 95.79                                   & 95.70                                        & 95.98                                       & 96.41                                         \\
\bottomrule
\end{tabular}
  \end{adjustbox}
  \label{tab:10}
\end{table*}

The data was also fed to MobileNet V3 Large models with base model resolutions of 512x384 pixels and 224x224 pixels, as we wanted to reduce the computation resources used. Eventually, the 22x224 pixels model achieved the highest accuracy on the TrashNet test set among the three models. The downscaling increased the accuracy. The 224x224 pixels model maintained the original test accuracy as the Efficient Net B0 benchmark model. It achieved a higher per-class precision on both the original test set and the new test set. This suggested that it was less biased towards different categories, which was expected for the application. \par

The larger input size does not always result in better accuracy. In \cite{Hesamian2015EffectNetwork}, the author demonstrated that the misclassification rate of the model reduced gradually when the resolution reduced from 2048x2048 pixels to 128x128 pixels. In our case, 224x224 might be a more suitable input size than 512x384, so the accuracy of MobileNet V3 Large increased when the input size was reduced. However, more experiments must be carried out to determine the best resolution for this application.\par

\begin{figure}[H]
 \centering
        \includegraphics[width=0.36\textwidth]{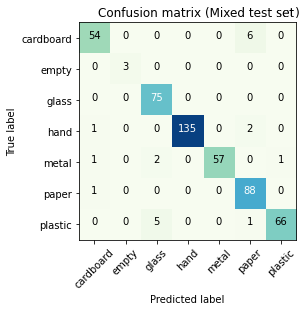}
        \caption{Confusion matrix of EffecientNet }
        \label{fig:8}
\end{figure}

\begin{figure}[H]
 \centering
        \includegraphics[width=0.36\textwidth]{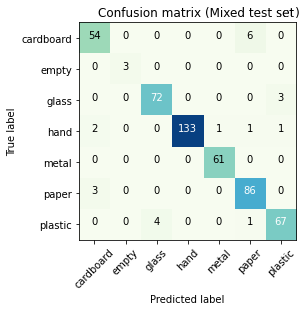}
        \caption{ Confusion matrix of MobileNet 224x224}
        \label{fig:9}
\end{figure}

The confusion matrix in Figure \ref{fig:8} and \ref{fig:9} shows that the model has the highest error rate in classifying between cardboard and paper. The MobileNet V3 model had the lowest sensitivity of 90\% on cardboard against paper, followed by the sensitivity of 94.37\% on plastic against glass. The EfficientNet and MobileNet model showed the same characteristics. 

Finally, The model was accelerated into the TensorRT model and ran on the Jetson Nano. Table \ref{tab:11} shows the inference speed of the three models. Compared to the 384x288 EfficientNet model, the MobileNet V3 computation speed increased because of the reduction of parameter size. The speed doubled for MobileNet V3 with a resolution of 224x224 pixels
\begin{table}[H]
  \centering
    \small
    \caption{Inference speed on Jetson Nano at maximum power mode}
  \begin{tabular}{p{3cm}p{2.5cm}p{2cm}}
    \toprule
    \cmidrule(r){1-3}
Model Architecture   & Input Resolution \newline (pixels) & Inference Per \newline second \\
    \midrule
EfficientNetB0       & 384x288                   & 19                   \\
MobileNet V3   Large & 512x384                   & 25                   \\
MobileNet V3   Large & 224x224                   & 40          \\                 
    \bottomrule
    
  \end{tabular}
  \label{tab:11}
\end{table}

To save power, we will run the models at 5 W power mode. This reduces the computation resources available and limits the inference speed. Table \ref{tab:12} shows that although the inference speed is reduced to 13 IPS for EfficentNet, the MobileNet model remains the same inference rate. The memory usage is reduced to 90.9\%. \par
\begin{table}[H]
  \centering
    \small
    \caption{Inference speed on Jetson Nano at 5 W power mode}
  \begin{tabular}{p{1.6cm}p{2cm}p{1cm}p{0.8cm}p{1cm}}
    \toprule
    \cmidrule(r){1-5}
Model \newline Architecture   & Base model Input Resolution (pixels) & Inference Per \newline second & Memory (\%) & Power \newline consumption (W)\\
    \midrule
EfficientNetB0       & 384x288                              & 13                   & 94.93       & 6.682                                                                  \\
MobileNet V3   Large & 224x224                              & 40                   & 90.90       & 4.698                                                                        \\                 
    \bottomrule
    
  \end{tabular}
  \label{tab:12}
\end{table}
An AI bin program is programmed using Python Tkinter library. The program interacts with the user through the graphics user interface (GUI) and calls the model to perform classifications. A demonstration video of the program can be found at \cite{XueyingLiDemonstrationPrototype}. The program occupied 6\% of Jetson Nano’s memory, so although it cannot run EfficientNet B0 model, MobileNet V3 large model worked fine for it. \par

Finally, as shown in Table \ref{tab:12}, the power consumption was reduced by 29.7\% when MobileNet V3 Large model was used. Running the GUI program did not increase the power consumption. The AI bin program consumed 3.97 W when the model was not inferencing, and Jetson Nano itself consumed 3 W when no program was running on it. Therefore no matter how light model of the model is, it will still consume 3 W power. Also, it took more than 20 minutes to load the model before inferencing, so Jetson Nano had to keep the program running at 3.97 W. \par

If a 92.5Wh battery pack is used, it will only last 23.3 hours in standby mode. The battery life will be shorter if the power consumption of the 10W bulb and the motors are taken into account. The battery pack must be changed manually every day if an independent battery source is used. However, the waste collection frequency in London is once a week.\par

So, the bin needs to connect to a cable or use a larger and more expensive power bank instead. This motivates us to test the performance of K210.\par

\subsection{Accuracy of the model trained with two datasets on K210}

\begin{figure}[H]
 \centering
        \includegraphics[width=0.5\textwidth]{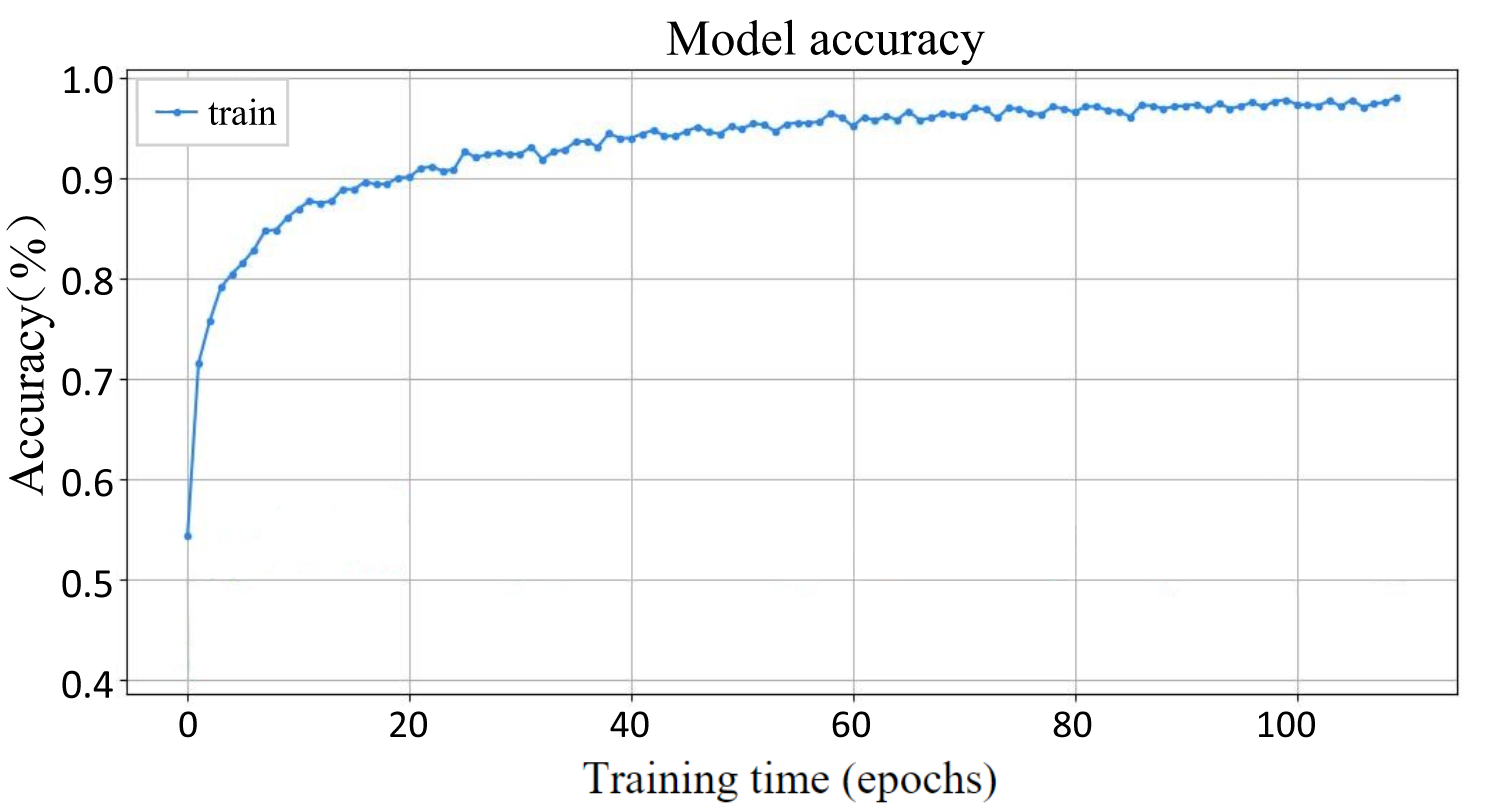}
        \caption{Change of Model accuracy during training}
        \label{fig:10}
\end{figure}

The accuracy of the K210 model was investigated in this section. The variation in training accuracy during the training process is shown in Figure \ref{fig:10}. 
At 50 epochs, the training accuracy was still increasing. After 100 epochs, it began to oscillate around 97\% and stopped rising further. So, the training was stopped at 100 epochs. \par

\begin{figure}[H]
 \centering
        \includegraphics[width=0.4\textwidth]{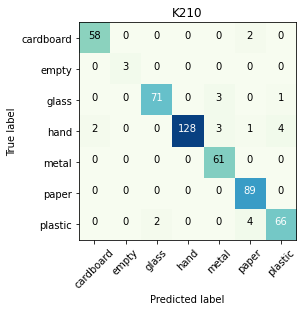}
        \caption{Confusion matrix of MobileNet V1 on K210}
        \label{fig:11}
\end{figure}

\begin{figure}
 \centering
        \includegraphics[width=0.4\textwidth]{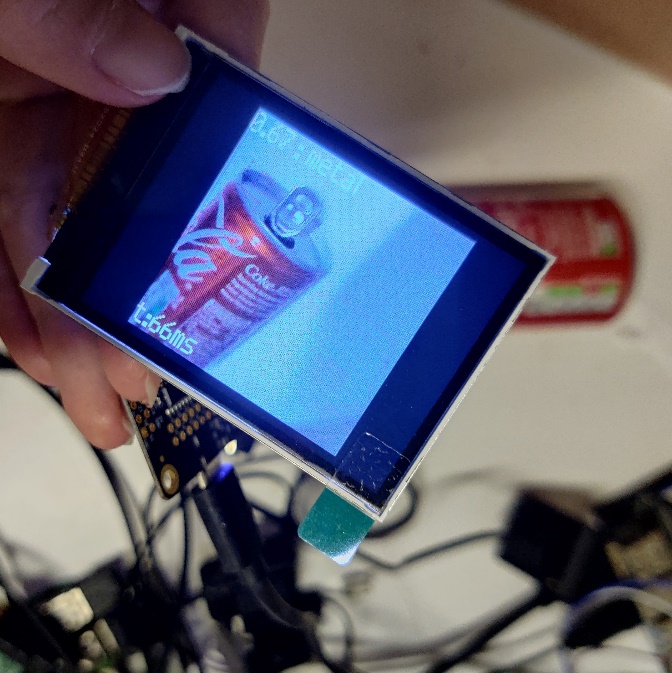}
        \caption{K210 is running the classification model}
        \label{fig:12}
\end{figure}

The model eventually achieved an accuracy of 96.64\% on the 5-class test set and 95.58\% on 7-class test set. The confusion matrix of the result is shown in Figure \ref{fig:11} . The accuracy of waste classification was higher than the previous models, while the 7-class accuracy was 0.5\% lower than the MobileNet V3 Large 224x224 model. This suggests that K210 could be a substitute for Jetson Nano on the classification task. 
Then, we loaded the model onto the K210 developer kit and ran the model. Figure \ref{fig:12} shows the photo of K210 when the model was running on it. The camera connected to it took photos of the object and the classification result is displayed on the top of the screen. The inference time was 66 ms, and the device only consumed 0.89W at interference. The power consumption was reduced by 80\% compared to the MobileNet model on Jetson Nano.
 
We only demonstrated K210's ability to run the classification model. In future research, the hyperparameters of the model can be fine-tuned to improve the model. Additionally, K210 also supports user interface programming with MircoPython, which could be used in the commercialized product.\par

\begin{figure}[H]
 \centering
        \includegraphics[width=0.5\textwidth]{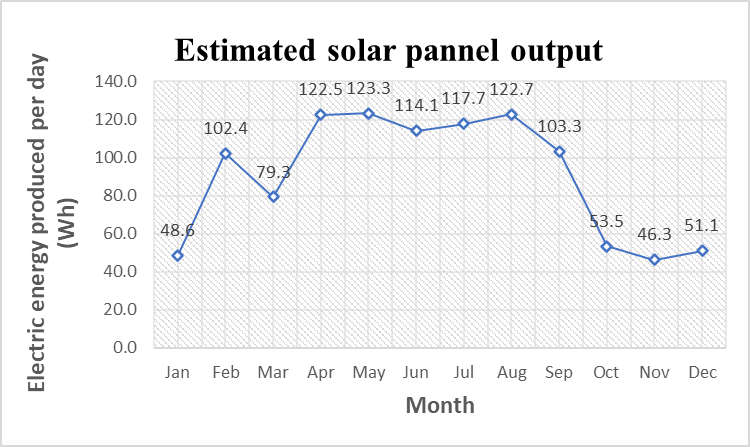}
        \caption{Estimated electric energy produced by solar panel in 2019}
        \label{fig:13}
\end{figure}

As the power consumption was less than 1 W, it can be powered by solar panels in the UK. Photovoltaic Geographical Information System \cite{EuropeanCommissionPHOTOVOLTAICSYSTEM} provided daily and monthly average solar radiation data in Europe and other areas through satellites. The monthly radiation on a plane normal to the sunrays in London in 2019 was downloaded from the database. The solar panel could be placed at the top of the bin, which will have at least 1600 $cm^{2}$ area. By using a solar panel with an efficiency of 22\% \cite{LixadaActivity}, the average energy produced a day in an ideal case can be calculated as:\cite{SaurNewsBureau2016HereSystem}
\begin{equation}
    \label{eq}
    E=A*r*H 
\end{equation}
where E is the maximum power output, A equals the solar panel area normal to the sunshine, r is the conversion efficiency, and H is the monthly irradiation on the plane normal to the sun rays. Figure \ref{fig:13} shows the electric energy produced by the solar panel per day.\par
As it can be observed, the solar panel will have the lowest output in November. Assuming that an ideal 48Wh battery is used to store the power energy produced by the solar panel and powers the K210 board, the battery can supply 1.9 W power in November ideally, while our application only takes 0.89 W. Additionally, it only takes 5 seconds to be switched on and starts detection immediately. This means it will not consume any power in standby mode, significantly reducing the requirements of energy consumption per day. So, using renewable energy to power it may be a feasible choice that can be explored in future studies. The combination of solar panels of 1600 $cm^{2}$ and 48 Wh board would typically cost between GBP 100 and GBP 200. It would be a cheaper option than Jetson Nano.

\section{Conclusion and future work}\label{sec:6}

This research aimed to build a real-time waste classification application on Jetson Nano with an accuracy above 95.38\% and a power consumption below 5 W. To eliminate the power supply cable connection and further encourage an eco-friendly charging method, K210 was proposed to replace Jetson Nano for the applications without self-improvement. \par

The application utilized MobileNet V3 Large as the primary classification model to reduce the model size and latency. We collected 1800 waste images in the target domain of this application with the detection box. The data was added to the TrashNet recycling waste dataset to improve the model's accuracy. Additionally, two new classes were inserted into the classification model that avoids trapping the user's hand and detects the cases when waste is not present. The model eventually achieved a 95.98\% accuracy on the test data and gave a throughput of 40 IPS at 4.698 W. Finally, we trained the MobileNet V1 model with the same training data, and it reached an accuracy of 96.64\%. The model was accelerated and ran on K210. It only consumed 0.89 W power. \par

The K210 developer board costs GBP 23, while the existing intelligent bin product costs USD 5200 \cite{Bin-e}. Therefore, considering the cost of solar batteries, conveyor belts, and physical shells, the production cost may still be attractive for commercialization. \par

Future works can continue to develop the trash bin with the two embedded systems. While Jetson Nano targets the highly accurate bin used in households and offices, K210 can be used in cheaper along-road self-powered bins. For Jetson Nano, the research can investigate the hyperparameters for continual training and implement the algorithm on Jetson Nano with limited computation resources that restricts the training time and samples. For K210, future works can build the circuit board for powering K210 with a solar panel. As the energy generated by the solar panel fluctuates depending on the sunshine, a power management system is required to ensure constant voltage output to K210. Moreover, the performance of the two applications could be tested with a larger unseen real-world test set with a size greater than 10\% of the training data. This will give a better estimation of their performance.\par

In conclusion, this paper has provided a state-of-the-art solution that reduced the power consumption of Jetson Nano by 30\% and increased the inference speed by 200\%. K210 only used 20\% of the power of Jetson Nano while maintaining accuracy. The use of this bin can increase the recycling rate up to 95\%, significantly improving the recycling rate in the UK. 
\section{Acknowledgements}
The authors would like to acknowledge their gratitude to Mindy Yang and Gary Thung who open-sourced the TrashNet dataset. The authors would also like to show gratitude to the Institute of Making staff for technical assistance with constructing the detection box.

%%%%%%%%%%%%%%   Bibliography   %%%%%%%%%%%%%%
\normalsize
\bibliographystyle{IEEEtran}
\bibliography{bibtex}
\end{document}